# Deep Model Compression: Distilling Knowledge from Noisy Teachers

Bharat Bhusan Sau    Vineeth N. Balasubramanian

Indian Institute of Technology, Hyderabad

{cs14mtech11002,vineethnb}@iith.ac.in

## Abstract

*The remarkable successes of deep learning models across various applications have resulted in the design of deeper networks that can solve complex problems. However, the increasing depth of such models also results in a higher storage and runtime complexity, which restricts the deployability of such very deep models on mobile and portable devices, which have limited storage and battery capacity. While many methods have been proposed for deep model compression in recent years, almost all of them have focused on reducing storage complexity. In this work, we extend the teacher-student framework for deep model compression, since it has the potential to address runtime and train time complexity too. We propose a simple methodology to include a noise-based regularizer while training the student from the teacher, which provides a healthy improvement in the performance of the student network. Our experiments on the CIFAR-10, SVHN and MNIST datasets show promising improvement, with the best performance on the CIFAR-10 dataset. We also conduct a comprehensive empirical evaluation of the proposed method under related settings on the CIFAR-10 dataset to show the promise of the proposed approach.*

*Keywords: Deep Learning, Model Compression, Teacher-Student Learning, Regularization, Noise*

## 1. Introduction

Since the remarkable success in object recognition by the famous work of Krizhevsky et al. [16] in 2012, deep learning has become very popular and replaced classical computer vision in a wide variety of real-world applications. We have observed performances on challenging real-world datasets that were once considered unachievable. With the introduction of networks like GoogLeNet [31], VGGNet [27] and ResNets [11] in recent times, networks are becoming deeper and deeper. Impressive progress has been made to train very deep networks with the invention of newer methods (as in [11]), as well as the relatively easier availability of computational resources today.

At the same time, consumer devices are also getting smaller and smaller in terms of mobility and portability, resulting in a widening chasm between deeper networks and their deployability on small mobile devices. Consumer devices have limited memory and computational capability to store or run such deep models (for e.g. the VGGNet network [27] requires 540 MB of storage, which is not suitable for a mobile device). Further, as described with relevant statistics by Han et al. in [10], running large networks on mobile devices increases memory access, which in turn consumes considerable battery power. This has resulted in a need for deep model compression methods.

Various deep model compression methods have been developed recently in the past 2-3 years to deal with these issues. These methods (described in Section 2) can be broadly categorized into parameter sharing methods, network pruning methods, 'dark knowledge' or teacher-student methods, and matrix decomposition methods. Although the performance of the compressed models have been good, most of these methods have focused on reducing only the storage complexity of the deep models. The compressed models have to be decompressed at runtime; the aforementioned issues of deployability on mobile devices thus continue to remain.

One approach among existing methods, which is not explored well enough, which can address this gap is the *teacher-student approach* ([1][3][12]) to deep model compression. This is a rather simple approach where a shallow student network is trained from a deep teacher network. However, considering that the student has to be a shallow network in order to minimize storage and runtime complexity, it becomes very challenging for the shallow student to achieve accuracy comparable to the deep teacher network. In this work, we propose a method based on the teacher-student framework to improve the performance of the student network, while remaining shallow. The impact of our contribution is directly relevant to the limitations discussed so far with existing deep model compression methods.

The key contributions of this work are as follows: (i) We focus our efforts on reducing storage as well as runtime complexity by building on the idea of teacher-student learning algorithm (considering a pretrained teacher such as AlexNet or VGGNet, this will also reduce training time complexity); (ii) While teacher-student learning algorithm is a promising approach for compressing deep models (in terms of storage, runtime and training time complexities), distilling knowledge from a single teacher can be limiting. We propose a simple noise-based regularization methodology to simulate learning from multiple teachers for better deep model compression; and (iii) We validate our methodology on three benchmark datasets: MNIST, SVHN and CIFAR-10 to highlight the usefulness of the proposed method. We also perform a comprehensive empirical analysis of the proposed method on CIFAR-10 to help identify success and failure cases which could help in adoption of the method to real-world applications.

We now present the related earlier work that is relevant to this work.

## 2. Background and Related Work

Model compression of deep networks has attracted attention very recently, and existing efforts can be broadly categorized into 4 kinds, each of which is described below.

**Parameter sharing methods:** Chen et al. [4] proposed a HashedNets model which used a low-cost hash function to group weights into hash buckets to share parameters where each hash bucket denotes a single parameter. Gong et al. [7] used $k$-means clustering to quantize the weights in fully connected layers and achieved upto 24x compression rate for their CNN network with only 1% loss on accuracy on the ImageNet challenge. Soulie et al. [28] used a regularization technique instead to coarsely quantize the weights of fully connected layers.

**Network pruning methods:** Han et al. [10] proposed to reduce the number of parameters by pruning weights which are below a threshold after the network is trained. They extended this work by using Huffman encoding [9] to reduce the number of parameters further. Leroux et al. [19] aimed at reducing computations by ignoring the convolutional filters which produce least activational strength. Srinivas and Babu [29] explored the redundancy among neurons, and proposed a data-free pruning methodology to remove redundant neurons.

**'Dark knowledge' methods:** The key idea of this group of methods, which is the focus of this work, is adopting a teacher-student strategy, where a large deep network trained for a given task teaches shallower student networks on the same task. The teacher-student training method was first proposed by Bucilu et al. [3] where they created synthetic data by labeling unlabeled data with a teacher model. This synthetically labeled data, which contains the knowledge captured by the teacher model, is then used to train a smaller student model. Ba and Caruana [1] proposed to train the student model by mimicking the logit values of the teacher model. This approach forms the baseline for this work, and we describe this method in more detail in Section 3.2. Romero et al. [24] extended this work by using intermediate hidden layer outputs as target values for training a student model. Hinton et al. [12] generalized this method by introducing a temperature variable in the softmax function. They showed that the softened outputs at higher temperatures convey important information. They termed this information which is expressed by the relative scores (probabilities) of the output classes as 'dark knowledge'.

**Matrix decomposition methods:** Denil et al. [5], Sainath et al. [25] and Nakkiran et al. [21] all adopted low-rank decomposition to compress the weights in different layers. Novikov et al. [23] converted the dense weight matrices of the fully connected layers to the Tensor Train format, such that the number of parameters is reduced by huge factor while preserving the expressive power of the layer. Other methods such as Denton et al. [6] that use matrix factorization attempt to speed up operations, for instance in the convolutional layer, but do not provide a holistic solution to the issues of complexity mentioned so far.

Compression of deep models is ideally important from all three perspectives: storage, train time and runtime complexities. Both parameter sharing and matrix decomposition methods concentrate on only the storage complexity of the deep models, but they fail to improve on runtime (or train time) complexity. Network pruning methods have shown very good performance on the storage complexity task, but are not aimed at reducing the runtime (or train time) complexity too. Teacher-student methods show promise in this direction of achieving compression across all complexities, but there has not been much follow-up work since it was proposed. This observation, supported by the potential of such methods in addressing the aforementioned issues, motivates our proposed work.

## 3. Proposed Methodology

Since our method is built on the teacher-student, viz. the 'dark knowledge' framework for deep model compression, we begin with a brief overview of teacher-student learning methods in Section 3.1, describe one such method in Section 3.2 (which we use as a baseline for comparison), and then provide our methodology in Section 3.3. We then show in Section 3.4 on how the proposed method is equivalent to noise-based regularization on the teacher.

### 3.1. Teacher-Student Learning

In teacher-student frameworks in deep learning, the teacher is a pretrained deep model, which is used to train another (typically shallow) model, called the student. There has been limited earlier work so far in this context, as described in Section 2. However, there are significant advantages of using a teacher-student framework beyond just model compression, as observed by Hinton et al. [12]:

- The 'dark knowledge' present in the teacher outputs works as a powerful target-cum-regularizer for the student model, as it provides soft targets that share helpful information.
- Convergence is typically faster than using only original 0/1 hard labels, due to the soft targets, that help training.
- A small amount of training data is generally sufficient to train the student network.

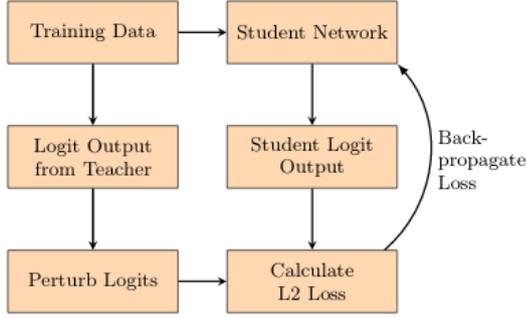

Figure 1. Training Shallow Students using the proposed Logit Perturbation Method

These advantages motivate us to extend this idea using noisy teachers. We now describe one of the teacher-student methods, which is used as a baseline for our experiments.

### 3.2. Student Learning using Logit Regression

Ba and Caruana [1] proposed a method to train the student directly on the log probability values $z$, also called *logits*, which is the output of the layer before softmax activation. The student network is trained in a regression setting using the logits, with its training data given by: $\{(x^{(1)}, z^{(1)}), ...., (x^{(i)}, z^{(i)}), ...., (x^{(n)}, z^{(n)})\}$. The L2 loss function minimized during training is given by:

$$L(x, z, \theta) = \frac{1}{2T} \sum_i \|g(x^{(i)}; \theta) - z^{(i)}\|_2^2 \qquad (1)$$

where:
- $T$ is the mini-batch size
- $x^{(i)}$ is $i^{th}$ training sample in the mini-batch
- $z^{(i)}$ is the corresponding logit output of the pretrained teacher for $x^{(i)}$
- $\theta$ is the set of student model parameters
- $g(x^{(i)}; \theta)$ is the student model's logit output for $x^{(i)}$

We build on this approach to develop our idea of noisy teachers.

### 3.3. 'Noisy Teachers': Student Learning using Logit Perturbation

The performance of shallow models in teacher-student framework settings was greatly improved by the methods proposed by Hinton et al. in [12], as well as Ba and Caruana in [1]. In this work, we explore this line of work further by asking the question: what if a student learns from multiple teachers? Analogous to human learning environments where a student can hone his understanding of a subject by learning the same topic from multiple experts of the field, we hypothesize that the performance of the student will improve under such a setting. However, instead of directly learning from multiple teachers, we propose a methodology to simulate the effect of multiple teachers, by injecting noise and perturbing the logit outputs of a teacher. The perturbed outputs, not only, simulate a multiple-teacher setting, but also results in noise in the loss layer, thus producing the effect of a regularizer. This new *noisy teacher* thus acts as a target-cum-regularizer and helps the student to learn better and produce results closer to the teacher network. We now describe the formulation of this method, which we also call *Logit Perturbation*.

Let $\xi$ be a vector of Gaussian noise with mean $\mu = 0$ and standard deviation $\sigma$. Dimension of $\xi$ is equal to the number of classes/logits in the teacher network. If $z^{(i)}$ is the logit layer output of the teacher model for $x^{(i)}$, then we modify $z^{(i)}$ as follows:

$$z'^{(i)} = (\mathbf{1} + \xi).z^{(i)} \qquad (2)$$

where $\mathbf{1}$ is a vector of ones and $i \in \mathbb{R}^n$ where $n$ is the number of classes. The loss function in Eqn 1 then becomes:

$$L(x, z', \theta) = \frac{1}{2T} \sum_i \|g(x^{(i)}; \theta) - z'^{(i)}\|_2^2 \qquad (3)$$

The value of $\sigma$ determines the amount of perturbation on the teacher's original logit values $z^{(i)}$. Higher the value of $\sigma$, higher is the perturbation. However, this perturbation need not be imposed on all samples. We instead select samples from the mini-batch with some fixed probability $\alpha$, and the logit values of the selected samples are then perturbed using Eqn 2.

Given a student model with initial parameters (weights) $\theta_0$, we then find its final parameters $\theta$ using Stochastic Gradient Descent (SGD), where in the $(t+1)^{th}$ iteration, $\theta$ is updated as follows:

$$\theta_{t+1} = \theta_t - \gamma_t . \sum_{(x,y) \in \mathcal{D}_t} \nabla_{\theta_t}[L(x, z', \theta)] \qquad (4)$$

Here, $T = |\mathcal{D}_t|$ is the size of the mini-batch randomly drawn from the training dataset $\mathcal{D}$, $\gamma_t$ is the learning rate, $L(x, z', \theta)$ is as in Eqn 3, and $\nabla_{\theta_t}[L(x, z', \theta)]$ is computed using gradient backpropagation.

**Algorithm 1** Training Student with Logit Perturbation
1: **Input:** `Training Data` $\mathcal{D}=(x,z)$, `probability` $\alpha$, `std` $\sigma$
2: **Initialization:** $\theta_0$ = Model Parameters of student model
3: **for** `each mini-batch` $\mathcal{D}_t = \{x_t, z_t\}$ **do**
4:    `Generate` $\xi$, $\xi \sim \mathcal{N}(\mathbf{0}, \sigma^2 \mathbf{I})$
5:    `Select samples from` $x_t$ `with probability` $\alpha$
6:    `Perturb corresponding logit values in` $z_t$ `using Eqn (2)`
7:    `Calculate L2 loss using Eqn (3)`
8:    `Update model parameters` $\theta_t$ `using Eqn (4)`
9: **end for**
10: **Output:** $\theta$ = Trained Student Model Parameters

In summary, some samples are selected with probability $\alpha$ from the mini-batch. The target logit values of the selected samples are perturbed using Eqn 2. The student network is then trained with the loss function in Eqn 3. A high-level flowchart of our method is shown in Figure 1. Algorithm 1 describes the pseudocode.

### 3.4. Equivalence to Noise-Based Regularization

It has been long established that noisy data in training helps to regularize a model (one of the earliest works showing this was [26]). Bishop showed in [2] that adding an L2 regularization term in the loss function is equivalent to adding Gaussian noise in the input data. The regularized loss function is given as:

$$L(x', \theta, z) = L(x, \theta, z) + R(\theta) \quad (5)$$

where $x'$ corresponds to $x$ with Gaussian noise, $L(x', \theta, z)$ is the L2 loss equivalent to noisy input data, $L(x, \theta, z)$ is the L2 loss for original input data, and $R(\theta)$ is the L2 regularizer. In our method, we perturb the target output $z$ with noise, instead of the input data $x$. It is now trivial to show that perturbing the target output $z$, the logit values of the teacher, is equivalent to adding a noise-based regularization term to the loss function. From Eqn 2, we get:

$$z'^{(i)} = (\mathbf{1} + \xi).z^{(i)} = z^{(i)} + \xi.z^{(i)}$$

Hence, we can rewrite the L2 loss in Eqn 3, as:

$$\begin{aligned} L(x, \theta, z') &= \|(z^{(i)} - g(x^{(i)}, \theta)) - \xi.z^{(i)}\|_2^2 \\ &= \|z^{(i)} - g(x^i, \theta)\|_2^2 + \|\xi.z^{(i)}\|_2^2 \\ &\quad + 2\|z^{(i)} - g(x^{(i)}, \theta)\|_2 * \|\xi.z\|_2 \\ &= L(x, \theta, z) + E_R \end{aligned}$$

where $E_R = \|\xi.z\|_2^2 + 2\|z^{(i)} - g(x^{(i)}, \theta)\|_2 \|\xi.z\|_2$ is the new regularizer that is based on the noise $\xi$. Thus, perturbing logits in the teacher network is equivalent to adding a noise-based regularizer to the loss function. (In Section 5, we show how this regularizer compares against using a standard L2 regularizer on the student network.) We now present our experimental results.

## 4. Experimental Results

We evaluated our method on three benchmark datasets: **MNIST** [18] and **SVHN** [22] for digit recognition, and **CIFAR-10** [15] for natural image recognition - each of which is described below. Stochastic Gradient Descent (SGD) is used to train all the networks used with a mini-batch size of 64. ADAM [14], which combines the ideas of momentum and adaptive learning rates, is used to adjust the learning rate in each iteration of SGD. Convergence was decided by testing on a hold-out validation set. For experiments in this section, we compare the performance of our method against the baseline performance of the teacher-student method as proposed by Ba and Caruana [1]. We also conducted preliminary experiments on comparing the proposed method against the work of Hinton et al. [12]; however, we found it non-trivial to identify the temperature at which their method's performance is maximized. In fact, we found it to give worse performance than [1] in our studies, and hence used [1] as the baseline comparison. Our results on all these datasets was promising, with the best improvement on the CIFAR-10 dataset. We further analyze the proposed method under various scenarios in Section 5.

### 4.1. MNIST

MNIST [18] is a popular dataset for handwritten digit recognition with 10 classes (0-9). The training set contains 50000 images and validation set contains 10000 images. All samples are $28 \times 28$ grayscale images. No pre-processing is done on the training data, keeping in line with earlier work on this dataset.

**Teacher Network:** We use a modified network of LeNet [17] as the teacher network on this dataset. LeNet has two convolutional layers and a fully connected layer followed by a 10-way classifier. We abbreviate the configuration of the teacher network as: [C5(S1P0)@20-MP2(S2)]-[C5(S1P0)@50-MP2(S2)]- FC500- FC10, where:

- C = Convolution Layer, with the following number indicating the size of the network, i.e. C5 indicates a convolutional layer with a $5 \times 5$ kernel
- S = Stride, i.e. S1 indicates a stride of 1
- P = Padding, i.e. P0 indicates a padding of zero
- @ = Number of kernels in Convolution Layer, i.e. @20 indicates 20 kernels in that layer
- MP = Max Pooling Layer, with the following number indicating the subsampling window, i.e. MP2 indicates max pooling of $2 \times 2$ windows
- FC = Fully Connected Layer, with the following number indicating the number of nodes in that layer

We use the above abbreviation for the sake of brevity in the remainder of this paper.

**Student Network:** We use a shallow network of two fully connected layers only with 800 neurons in each layer, as described in Hinton et al. [12], as our student network. The architecture can be encoded as: FC800-FC800-FC10.

**Results:** The teacher model achieved 68 test errors (out of 10000 test samples, error rate = 0.0068). The student model achieved 97 test errors (error rate = 0.0097) with the baseline teacher-student method (logit regression method described in Section 3.2). As the difference of performance between teacher and student is not very high, we chose the sample selection probability $\alpha = 0.15$, i.e., around 15% samples of each mini-batch are selected for logit perturbation. The perturbation was done at various noise levels (Gaussian with $\mu = 0$, different $\sigma$s) as shown in Table 1. This noise is added directly to the unnormalized logits in all our experiments in this work. We see that there is consistent improvement of performance of the student while applying perturbation to logits.

Table 1. Results on MNIST (Baseline error rate = 0.0097)

| Noise Level($\sigma$=std) | Error Rate | % Improvement |
|---|---|---|
| 0.10 | 0.0096 | 1.0% |
| 0.20 | 0.0093 | 4.1% |
| 0.30 | 0.0094 | 3.1% |
| 0.40 | 0.0087 | 10.3% |
| 0.50 | 0.0087 | 10.3% |
| 0.60 | 0.0090 | 7.2% |
| 0.70 | 0.0090 | 7.2% |
| 0.80 | 0.0086 | 11.3% |
| 0.90 | 0.0086 | 11.3% |
| 1.00 | 0.0087 | 10.3% |

### 4.2. SVHN

The Street View House Numbers (SVHN) [22] is a real-world image dataset containing cropped digits in house numbers from Google Street View images. It contains a much higher number of training and testing samples than MNIST. It has 73257 training samples, 26032 test samples, along with 531131 additional samples which can be used for training, each of which is an RGB image of size $32 \times 32$. As in earlier work [8] that have used this dataset, we selected 400 samples per category from the training set and 200 samples per category from the additional set to constitute a validation set of 6000 samples (which is used to decide if the training has converged sufficiently). The remaining 598388 samples are combined to form the training set. We also preprocessed all the data using local contrast normalization as in [13].

**Teacher Network:** We used the Network-in-Network [20] model as our teacher network for this dataset. It has several convolutional layers, and its uniqueness lies in replacing the fully connected layer with a global average pooling layer (denoted by AP, where the number following AP denotes the subsampling window size). The full configuration of the network can be written as: [C5(S1P2)@192]-[C1(S1P0)@160]- [C1(S1P0)@96-MP3(S2)]- D0.5-[C5(S1P2)@192]- [C1(S1P0)@192]- [C1(S1P0)@192-AP3(S2)]- D0.5- [C3(S1P1)@192]- [C1(S1P0)@192]-[C1(S1P0)@10]- AP8(S1). D0.5 denotes DropOut in between the corresponding layers with a probability of 0.5.

**Student Network:** We used a version of **LeNet**, as our student network. The network can be written as: [C5(S1P2)@32-MP3(S2)]- [C5(S1P2)@64-MP3(S2)]-FC1024-FC10. We decided to have two convolutional layers in our student architecture based on the result shown by Urban et al. [32]. They showed that in order to mimic a large convnet and achieve significant performance, the student layer must have some convolutional layers.

**Results:** The teacher model gave 3.82% error on the SVHN test set. The baseline student model (trained, as before, using the logit regression method described in Section 3.2) gave a 4.6% error rate. Once again, as the difference between student's and teacher's performance is not very high, we selected $\alpha = 0.15$. Using the proposed logit perturbation methodology, our student model achieved a 4.45% error rate within the range of standard deviations considered for the noise. Detailed results are shown in Table 2. We also observe that higher noise levels result in deterioration of performance on this dataset, indicating the importance of setting the values of $\alpha$ and $\sigma$ appropriately.

Table 2. Results on SVHN (Baseline error rate = 4.6%)

| Noise Level($\sigma$=std) | Error Rate(%) | % Improvement |
|---|---|---|
| 0.10 | 4.51 | 2.0% |
| 0.20 | 4.46 | 3.0% |
| 0.30 | 4.45 | 3.3% |
| 0.40 | 4.54 | 1.3% |
| 0.50 | 4.57 | 0.7% |
| 0.60 | 4.51 | 2.0% |
| 0.70 | 4.61 | -0.2% |
| 0.80 | 4.68 | -1.7% |
| 0.90 | 4.72 | -2.6% |
| 1.00 | 4.81 | -4.6% |

### 4.3. CIFAR-10

CIFAR-10 [15] is a popular dataset for small-scale image recognition. The dataset contains 10 classes of natural images with a total of 50000 training and 10000 testing samples, each of which is a $32 \times 32$ RGB image. We preprocessed the data by subtracting per-pixel mean and also took mirror image of the samples for data augmentation during training (thus increasing the training set to 100000 images).

**Teacher Network:** We used the same teacher network (the Network-in-Network model) as used for the SVHN dataset (Section 4.2).

**Student Network:** The student network we used here is also a modified version of the LeNet architecture, with two convolutional layers and one fully connected layer. The architecture of our student network for this dataset can be given as: [C5(S1P2)@64-MP2(S2)]- [C5(S1P2)@128-MP2(S2)]-FC1024-FC10.

**Results:** Trained on 100000 data samples, the teacher network obtained 8.4% error rate. The baseline student model (trained using the logit regression method described in Section 3.2) provided a best performance of a 21.94% error rate. The results of the proposed method are described in Table 3. We observe that the proposed method gives a healthy improvement over the baseline method. Considering the large difference between teacher and student networks' performances, we chose $\alpha = 0.5$ for this dataset. This indicates that a good amount of perturbation helps the student network achieve better performance in this case.

Table 3. Results on CIFAR-10 (Baseline error rate = 21.94%)

| Noise Level($\sigma$=std) | Error Rate (%) | % Improvement |
|---|---|---|
| 0.10 | 21.04 | 4.1% |
| 0.20 | 20.28 | 7.6% |
| 0.30 | 20.09 | 8.4% |
| 0.40 | 19.79 | 9.8% |
| 0.50 | 19.27 | 12.2% |
| 0.60 | 19.15 | 12.7% |
| 0.70 | 19.39 | 11.6% |
| 0.80 | 18.72 | 14.7% |
| 0.90 | 18.68 | 14.9% |
| 1.00 | 18.96 | 13.6% |

The aforementioned results corroborate the promise of the proposed methodology. We present a detailed analysis of the method across different settings, including runtime complexities, in the subsequent section.

## 5. Discussions and Analysis

In this section, we perform a comprehensive analysis of the performance of the proposed method under different conditions: varying noise in the teacher, comparison of a noisy teacher vs a student regularized directly with noise, the equivalence of this noisy teacher framework to learning from multiple teachers, as well as the comparison of the proposed noisy teacher approach to other regularization techniques. Our experimental results in this section are all carried out on the CIFAR-10 dataset.

### 5.1. Varying Noise in the Teacher

In Section 4.3, we selected samples from a mini-batch with a fixed probability $\alpha = 0.5$ and varied the $\sigma$ to experiment on different noise levels. In this section, we fix $\sigma$ (standard deviation of the Gaussian noise), and vary the probability $\alpha$ of the number of logit values on which noise is added. The results are shown in Table 4. For this test, we kept $\sigma = 0.6$ constant for all the experiments. The teacher and student networks are the same as described in Section 4.3. From the table, we see that higher values of $\alpha$ helped the student to achieve better performance. The performance reaches the best at $\alpha = 0.8$, i.e, when around 80% samples of mini-batch are selected for perturbation.

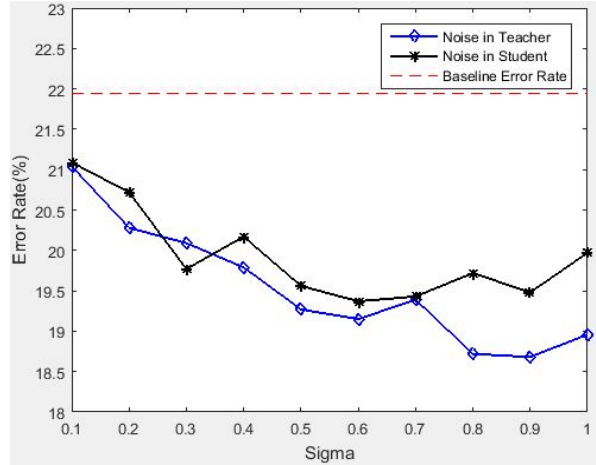

Figure 2. CIFAR-10: noise in teacher vs noise in student (Section 5.2). Base error rate refers to the baseline error rate obtained using the logit regression method of training the student network. Sigma refers to the standard deviation of the noise added.

Our empirical studies revealed that the value of $\alpha$ for which a student network may give best result depends on the performance gap between the baseline student and teacher network. If this gap is large, a higher amount of noise is required to help the student train better and achieve its best possible performance. If this gap is small, as in the case of MNIST and SVHN (Sections 4.1 and 4.2), a lower number of noisy logits is more helpful.

Table 4. Varying Number of Noisy Logits, CIFAR-10 (Baseline error rate = 21.94%)

| $\alpha$ | Error Rate (%) | % Improvement |
|---|---|---|
| 0.1 | 20.14 | 8.2% |
| 0.2 | 19.73 | 10.1% |
| 0.3 | 19.39 | 11.6% |
| 0.4 | 19.42 | 11.5% |
| 0.5 | 19.15 | 12.7% |
| 0.6 | 19.05 | 13.2% |
| 0.7 | 19.04 | 13.2% |
| 0.8 | 18.93 | 13.7% |
| 0.9 | 19.00 | 13.4% |
| 1.0 | 19.00 | 13.4% |

### 5.2. Noise in Teacher vs Noise in Student

In the experiments in Section 4.3, we saw that noise in the teacher, i.e., perturbing the logit outputs of the teacher network, has a positive regularization effect on the student, and helps it to achieve higher performance than the baseline (only logit regression). In this section, we ask the question if such an improvement in performance can be obtained by using the baseline method directly and instead regularizing (using weight decay) the student network during its training. This can be considered as directly adding noise to the student while training. The key difference between these two settings is that teacher network is already trained whereas student network is being trained. Hence, the logit output from a noisy teacher for a sample remains the same throughout the training phase whereas in case of a noisy student, this keeps changing. In this experiment, we modified the logit output from student in the same way we did for the teacher logit outputs. We used the same teacher and student network as in Section 4.3. The results are shown in Figure 2. The figure shows that a noisy teacher is more helpful than a noisy student. We believe this could be attributed to the understanding that learning from a noisy teacher with random noise is equivalent to learning from multiple similar teachers with minor variations though. Drawing a parallel to human learning environments, it would only be expected that students learn well under such conditions.

### 5.3. Comparison with DropOut

While adding noise to the student is one form of regularization (as discussed in Section 3.4), we also compared the performance of the proposed method against another often used regularization method: DropOut [30]. We used the same teacher and student network described in Section 4.3, with one DropOut layer added after the fully connected layer of the student network. We studied the performance of the student under different DropOut ratios (i.e. varying probabilities of dropping nodes). The results are summarized in Table 5. We notice that DropOut does not perform as well as the proposed *noisy teacher regularizer*. The best improvement over the baseline that DropOut achieved was a decrease in error rate by 1.53%, while the proposed method performed significantly better (as in Table 3) with a best decrease in error rate of 3.26%.

Table 5. Comparison with Dropout (Baseline error rate without regularization = 21.94%; Best performance of student trained with proposed method = 18.68%)

| Dropout Ratio | Error Rate(%) |
|---|---|
| 0.10 | 26.48 |
| 0.20 | 25.34 |
| 0.30 | 24.11 |
| 0.40 | 23.25 |
| 0.50 | 22.46 |
| 0.60 | 21.15 |
| 0.70 | 20.95 |
| 0.80 | 20.41 |
| 0.90 | 22.18 |

### 5.4. More Results

In this section, we describe other experiments that we carried out to study related issues: (i) the impact of random noise level in each iteration of training; and (ii) learning from multiple teachers.

**Effect of Random Noise Levels:** As described in Section 3, we used a Gaussian noise to perturb the logit output of the teacher in this work. In all previous experiments, we used a fixed $\sigma$ (standard deviation, or noise level, as we called it) in all iterations. So, the total perturbation is, in a sense, controlled. To study the impact of more *uncontrolled* noise across mini-batches of training, we investigate the effect of using a random value of $\sigma$ in each mini-batch iteration, which causes the perturbation to vary across iterations. We select a value of $\sigma$ uniformly in $[0.01, 1]$, and accordingly generate Gaussian noise to perturb the teacher outputs. The teacher and student network remain the same as in Section 4.3. We observed an improvement (decrease) in error rate of 2.86% over the baseline reported in Section 4.3. While this is lesser than the best performance we got using a fixed $\sigma$ (a best decrease in error rate of 3.26% when $\sigma = 0.9$ in Table 3), the healthy improvement in performance still points to the conclusion that even when there is limited budget/bandwidth for experimenting with various noise levels, using random noise levels can be beneficial in training the student network.

**Learning from Multiple Teachers:** In Section 3, we described the proposed logit perturbation method to be analogous to learning from multiple teachers, where the noise dictates the amount of diversity between the teachers. In this section, we extend this perspective to this work by experimenting with two teachers on the same student to study their effect on the student's performance. The two teachers we use are: (i) Network-in-Network model (Teacher1), as described in Section 4.3 and (ii) A modified version of Alexnet [16](Teacher2). The configuration of Teacher2 is: [C5(S1P2)@96-MP3(S2)]- [C5(S1P2)@256-MP3(S2)]- [C3(S1P1)@384]- [C3(S1P1)@384]- [C3(S1P1)@256-MP3(S2)]- FC2048-D0.5- FC2048- D0.5- FC10. The student network(modified LeNet) is the same as described in Section 4.3.

The baseline error rate that we obtained in our experiments of Teacher1 was 8.4%, and that of Teacher2 was 13.99%. To let the student learn from two teachers via the baseline logit regression method (We did not use noisy teachers here, since the objective is to study learning from multiple teachers as is; we used the baseline algorithm as in Section 3.2), we took the geometric mean of the logits from the two teacher networks and used that as the target logit value for the student. We obtained a best performance (error rate) of 20.44% for the student trained in this manner, whereas the same same student trained using only Teacher1 gave 21.94% error rate (as in Section 4.3), and using only Teacher2 gave 22.62%. This supports our hypothesis that learning from multiple teachers (of which noisy teachers are a special case) results in a positive effect on the student.

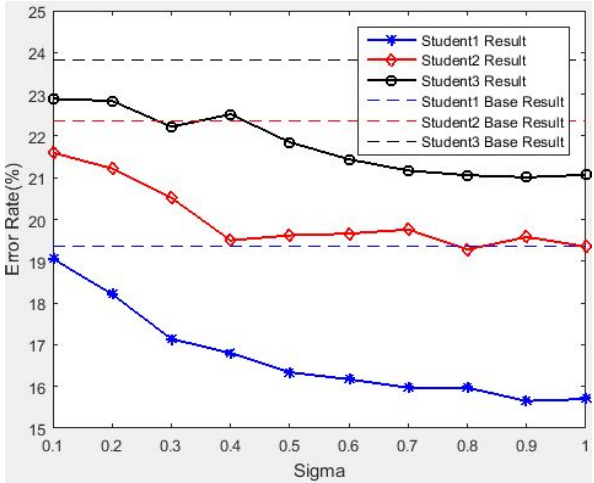

Figure 3. Performance of shallow models at different compression ratios (Section 5.5).

### 5.5. Runtime Compression of Shallow Students

Considering that the key application of the proposed idea in deep learning applications is in model compression, we study the issue of compression in this section. Recent methods, especially in the last 2 years, have done very well in reducing storage complexity (e.g. [9]). However, the compressed networks, in most such cases, need to be *decompressed* before deployment in real-time, thus resulting in a higher runtime complexity. Evidently, using shallow student networks has the potential to compress deep networks from both perspectives of storage and runtime complexity. The reduced storage complexity of shallow student networks is evident due to the reduced depth (similarly, reduced train-time complexity is also evident given a pre-trained teacher). In particular, we study the runtime complexity of shallow student networks in this section.

We build three shallow networks of depth 4[1], which require significantly lower computations than standard deep networks used today, for the CIFAR-10 dataset. We study the performance of these shallow networks using the logit regression method (standard teacher) and proposed logit perturbation method (noisy teacher) along with their runtime compression performance.

**Teacher Network**: We use the same teacher as described in Section 4.3, which has an error rate of 8.4% (or accuracy of 91.6%) on CIFAR-10. This network requires 223 million multiplications in each forward pass (for convenience of analysis, without loss in generality, we consider multiplications as our measure of computations).

**Student1:** Its architecture is: [C5(S1P2)@64-MP2(S2)]- [C5(S1P2)@112-MP2(S2)]- [C3(S1P1)@128-MP2(S2)]- FC1024- FC10. Total computations required for this student network is 61 million/forward pass. The computations in each layers are: 5-46-8-2 million/forward pass. Hence, runtime compression obtained w.r.t the teacher = 223/61 = 3.66.

[1]We used student networks of depth 3 until now, and use student networks of depth 4 in this section to highlight the issue of compression.

**Student2:** The architecture of this student is: [C5(S1P2)@32-MP2(S2)]- [C5(S1P2)@32-MP2(S2)]- [C3(S1P1)@64-MP2(S2)]- FC1024- FC10. Total computations required for this network are 11.2 million/forward pass. Layerwise computations are: 2.5-6.5-1.2-1 million/forward pass. Hence, its runtime compression = 223/11.2 = 19.91.

**Student3:** In this student, we reduced the number of filters in the first layer drastically. The architecture is: [C5(S1P2)@16-MP2(S2)]- [C5(S1P2)@32-MP2(S2)]- [C3(S1P1)@64-MP2(S2)]- FC1024- FC10. Hence, the number of computations in each layer are: 1.25-3.25-1.2-1 million/forward pass, with the total computations hence reduced to 6.7 million/forward pass. The run-time compression of this network w.r.t. teacher = 223/6.7 = 33.28.

We compared the performance obtained from our method and the baseline (logit regression) method for these student networks on CIFAR-10. For all the student networks trained using the proposed method, $\alpha$ was set to 0.5, and the noise levels were varied between 0.1 and 1.0, and the best result was chosen. The results are shown in Figure 3 and Table 6. Evidently, as the amount of computations decrease, the performance also starts to decrease. Yet in all cases, the proposed method provides superior performance to the baseline logit regression method. While shallow networks expectedly do not perform as well as the original deep teacher network, we highlight that the use of noisy teachers while training shallow student networks provide a significant boost to performance, while maintaining high levels of compression both in terms of storage and runtime.

Table 6. Error Rate(%) on CIFAR-10 with student networks of varying compression ratios (Logit Regression = Baseline standard teacher; Our Method = Noisy teacher)

| Student | Compression Ratio | Logit Regression | Our Method |
|---------|-------------------|------------------|------------|
| Student1 | 3.66 | 19.36 | 15.65 |
| Student2 | 19.91 | 22.36 | 19.28 |
| Student3 | 33.28 | 23.83 | 21.01 |

### 6. Conclusions

In this work, we presented a method based on teacher-student learning framework for deep model compression, considering both storage and runtime complexities. Our noise-based regularization method helped the shallow student model to do significantly better than the baseline teacher-student algorithm. The proposed method can be viewed also as simulating learning from multiple teachers, thus helping student models to get closer to the teacher's performance.

# References

[1] J. Ba and R. Caruana. Do deep nets really need to be deep? In *Advances in neural information processing systems*, pages 2654–2662, 2014.

[2] C. M. Bishop. Training with noise is equivalent to tikhonov regularization. *Neural computation*, 7(1):108–116, 1995.

[3] C. Bucilu, R. Caruana, and A. Niculescu-Mizil. Model compression. In *Proceedings of the 12th ACM SIGKDD international conference on Knowledge discovery and data mining*, pages 535–541. ACM, 2006.

[4] W. Chen, J. T. Wilson, S. Tyree, K. Q. Weinberger, and Y. Chen. Compressing neural networks with the hashing trick. *CoRR, abs/1504.04788*, 2015.

[5] M. Denil, B. Shakibi, L. Dinh, N. de Freitas, et al. Predicting parameters in deep learning. In *Advances in Neural Information Processing Systems*, pages 2148–2156, 2013.

[6] E. L. Denton, W. Zaremba, J. Bruna, Y. LeCun, and R. Fergus. Exploiting linear structure within convolutional networks for efficient evaluation. In *Advances in Neural Information Processing Systems*, pages 1269–1277, 2014.

[7] Y. Gong, L. Liu, M. Yang, and L. Bourdev. Compressing deep convolutional networks using vector quantization. *arXiv preprint arXiv:1412.6115*, 2014.

[8] I. J. Goodfellow, D. Warde-Farley, M. Mirza, A. C. Courville, and Y. Bengio. Maxout networks. *ICML (3)*, 28:1319–1327, 2013.

[9] S. Han, H. Mao, and W. J. Dally. Deep compression: Compressing deep neural network with pruning, trained quantization and huffman coding. *CoRR, abs/1510.00149*, 2, 2015.

[10] S. Han, J. Pool, J. Tran, and W. Dally. Learning both weights and connections for efficient neural network. In *Advances in Neural Information Processing Systems*, pages 1135–1143, 2015.

[11] K. He, X. Zhang, S. Ren, and J. Sun. Deep residual learning for image recognition. *arXiv preprint arXiv:1512.03385*, 2015.

[12] G. Hinton, O. Vinyals, and J. Dean. Distilling the knowledge in a neural network. *arXiv preprint arXiv:1503.02531*, 2015.

[13] K. Jarrett, K. Kavukcuoglu, Y. Lecun, et al. What is the best multi-stage architecture for object recognition? In *2009 IEEE 12th International Conference on Computer Vision*, pages 2146–2153. IEEE, 2009.

[14] D. Kingma and J. Ba. Adam: A method for stochastic optimization. *arXiv preprint arXiv:1412.6980*, 2014.

[15] A. Krizhevsky and G. Hinton. Learning multiple layers of features from tiny images. 2009.

[16] A. Krizhevsky, I. Sutskever, and G. E. Hinton. Imagenet classification with deep convolutional neural networks. In *Advances in neural information processing systems*, pages 1097–1105, 2012.

[17] B. B. Le Cun, J. S. Denker, D. Henderson, R. E. Howard, W. Hubbard, and L. D. Jackel. Handwritten digit recognition with a back-propagation network. In *Advances in neural information processing systems*. Citeseer, 1990.

[18] Y. LeCun, L. Bottou, Y. Bengio, and P. Haffner. Gradient-based learning applied to document recognition. *Proceedings of the IEEE*, 86(11):2278–2324, 1998.

[19] S. Leroux, S. Bohez, C. De Boom, E. De Coninck, T. Verbelen, B. Vankeirsbilck, P. Simoens, and B. Dhoedt. Lazy evaluation of convolutional filters. *arXiv preprint arXiv:1605.08543*, 2016.

[20] M. Lin, Q. Chen, and S. Yan. Network in network. *arXiv preprint arXiv:1312.4400*, 2013.

[21] P. Nakkiran, R. Alvarez, R. Prabhavalkar, and C. Parada. Compressing deep neural networks using a rank-constrained topology. 2015.

[22] Y. Netzer, T. Wang, A. Coates, A. Bissacco, B. Wu, and A. Y. Ng. Reading digits in natural images with unsupervised feature learning. 2011.

[23] A. Novikov, D. Podoprikhin, A. Osokin, and D. P. Vetrov. Tensorizing neural networks. In *Advances in Neural Information Processing Systems*, pages 442–450, 2015.

[24] A. Romero, N. Ballas, S. E. Kahou, A. Chassang, C. Gatta, and Y. Bengio. Fitnets: Hints for thin deep nets. *arXiv preprint arXiv:1412.6550*, 2014.

[25] T. N. Sainath, B. Kingsbury, V. Sindhwani, E. Arisoy, and B. Ramabhadran. Low-rank matrix factorization for deep neural network training with high-dimensional output targets. In *2013 IEEE International Conference on Acoustics, Speech and Signal Processing*, pages 6655–6659. IEEE, 2013.

[26] J. Sietsma and R. J. Dow. Creating artificial neural networks that generalize. *Neural networks*, 4(1):67–79, 1991.

[27] K. Simonyan and A. Zisserman. Very deep convolutional networks for large-scale image recognition. *arXiv preprint arXiv:1409.1556*, 2014.

[28] G. Soulié, V. Gripon, and M. Robert. Compression of deep neural networks on the fly. *arXiv preprint arXiv:1509.08745*, 2015.

[29] S. Srinivas and R. V. Babu. Data-free parameter pruning for deep neural networks. *arXiv preprint arXiv:1507.06149*, 2015.

[30] N. Srivastava, G. E. Hinton, A. Krizhevsky, I. Sutskever, and R. Salakhutdinov. Dropout: a simple way to prevent neural networks from overfitting. *Journal of Machine Learning Research*, 15(1):1929–1958, 2014.

[31] C. Szegedy, W. Liu, Y. Jia, P. Sermanet, S. Reed, D. Anguelov, D. Erhan, V. Vanhoucke, and A. Rabinovich. Going deeper with convolutions. In *Proceedings of the IEEE Conference on Computer Vision and Pattern Recognition*, pages 1–9, 2015.

[32] G. Urban, K. J. Geras, S. E. Kahou, O. Aslan, S. Wang, R. Caruana, A. Mohamed, M. Philipose, and M. Richardson. Do deep convolutional nets really need to be deep (or even convolutional)? *arXiv preprint arXiv:1603.05691*, 2016.